\documentclass{article}

\PassOptionsToPackage{table}{xcolor}
\usepackage{iclr2027_conference,times}
\iclrfinalcopy

\usepackage[utf8]{inputenc}
\usepackage[T1]{fontenc}
\usepackage{url}
\usepackage{booktabs}
\usepackage{amsmath}
\usepackage{amssymb}
\usepackage{amsfonts}
\usepackage{graphicx}
\usepackage{microtype}
\usepackage[table]{xcolor}
\usepackage{float}
\usepackage{subcaption}
\usepackage{circledsteps}
\pgfkeys{/csteps/fill color=black, /csteps/inner color=white}
\usepackage{placeins}
\usepackage{needspace}
\usepackage{wrapfig}
\usepackage{tikz}
\usetikzlibrary{arrows.meta,positioning,calc}
\definecolor{dealorange}{HTML}{D55E00}
\definecolor{dealgreen}{HTML}{009E73}
\usepackage{multirow}
\usepackage{algorithm}
\usepackage{algpseudocode}
\usepackage{enumitem}
\usepackage{tcolorbox}
\usepackage{hyperref}

\title{DEALS: \underline{D}ecentralized \underline{E}xpertise-\underline{A}ware \underline{L}oad \underline{S}erving for Multi-Agent LLM Systems}
\author{Jingjuan Huang$^{1}$, Wenbin Wang$^{1}$, Yanchuan Yin$^{2}$, Alvaro Velasquez$^{2}$, Jia Liu$^{1}$ \\
$^{1}$Ohio State University \qquad $^{2}$University of Colorado at Boulder}

\begin{document}
\maketitle
\lhead{Preprint}

\begin{abstract}
Multi-agent systems (MAS) have recently emerged as an effective approach for coordinating large language model (LLM)-based agents to solve complex tasks through structured interactions.
In practice, MASs often handle a stream of heterogeneous and complex tasks, requiring agents to decompose each task and then self-organize and self-evolve to adapt to incoming tasks while sharing execution resources.
However, most early approaches to MASs rely on centralized controllers or fixed coordination patterns, which can limit scalability or adaptability. 
In contrast, existing decentralized and dynamic MASs often require training dedicated routers or invoking LLMs for agent selection, resulting in substantial computational costs and coordination overhead. 
To address these challenges and enable efficient task-level self-organization and self-evolution for task- and workload-level collaboration, we propose \underline{D}ecentralized \underline{E}xpertise-\underline{A}ware \underline{L}oad \underline{S}erving (DEALS), a decentralized and low-complexity framework that enables agents to self-organize and dynamically route concurrent tasks for processing.
Specifically, each agent maintains local queues of incoming tasks, and its router decides whether to process a task locally or forward it to a neighbor based on differences in backlog and success rate.
Meanwhile, executors process independent tasks concurrently within and across agents, and partially solved tasks can be resumed by other agents.
Experiments show that DEALS not only improves performance along multiple dimensions (e.g., answer accuracy and task throughput) in both homogeneous and heterogeneous agent pools, but also balances agent expertise and workload in a self-organized manner, enabling effective decentralized coordination.
\end{abstract}

\section{Introduction}
\label{sec:introduction}

In recent years, large language model (LLM)-based agents have demonstrated astonishing capabilities in reasoning and tool use \citep{brown2020language,yao2023react}.
However, as agentic tasks grow increasingly complex, a single LLM-based agent may struggle to reliably solve complex problems and sustain long-horizon executions \citep{liu2024lost,liu2024agentbench}.
To address this challenge, multi-agent systems (MAS) have recently emerged as an effective approach for coordinating multiple LLM-based agents to solve complex agentic tasks through structured interactions, such as decomposing a complex tasks, assign subtasks to agents with specialized roles, and collaboratively producing solutions \citep{li2023camel,wu2024autogen,hong2024metagpt}.

In the literature, early approaches to MASs relied on centralized controllers or hand-designed static role specializations with fixed interaction patterns~\citep{li2023camel,hong2024metagpt}. 
Later methods focused on optimizing collaboration in terms of graph connectivity~\citep{zhuge2024gptswarm,zhang2025agentprune}, workflow structures~\citep{zhang2025aflow}, and team composition~\citep{liu2024dylan}. 
However, these approaches optimize collaboration statically and therefore still lack the capacity for self-organization and self-evolution.
More recently, researchers have started to consider self-organization and self-evolution for MASs through decentralized routing decisions \citep{yang2025agentnet} and private memories \citep{hao2026decentmem}.
Across these approaches, the primary focus is on how agents collaborate to solve an individual task. Moreover, emergent self-organization and self-evolution behaviors are typically induced through reinforcement learning techniques~\citep{zhang2025landscape}, which can incur substantial training costs and suffer from stability issues. 

In practice, however, MASs often need to serve a continuous stream of heterogeneous tasks arriving concurrently and/or sequentially~\citep{lin2024parrot,luo2025autellix}. Supporting multiple heterogeneous tasks simultaneously not only introduces more complex challenges in self-organization and self-evolution, but can also lead to workload imbalance across agents if not carefully managed, thereby degrading system throughput and increasing latency.
To date, most existing work on multi-task routing in MASs relies on centralized coordination~\citep{ni2026chimera,da2026routebalance}. However, centralized coordination can become a scalability bottleneck and a single point of failure as the system grows. 
Moreover, some methods require separately trained routers~\citep{yue2025masrouter}, while others rely on LLM inference for agent selection~\citep{wu2024autogen}. 
These approaches can incur substantial training costs, consume significant inference resources, and introduce additional coordination latency. Motivated by these limitations, we ask the following research question:

\begin{tcolorbox}[left=1.2pt,right=1.2pt,top=1.2pt,bottom=1.2pt,before skip=4pt,after skip=4pt]
\textbf{(Q)}: Can we design a self-organizing and self-evolving MAS with low-complexity decentralized routing to collaboratively solve multiple concurrent tasks with high accuracy and low latency?
\end{tcolorbox}

In this paper, we answer this question affirmatively by proposing the \underline{D}ecentralized \underline{E}xpertise-\underline{A}ware \underline{L}oad \underline{S}erving method (DEALS), a low-complexity decentralized framework that facilitates self-organization and self-evolution in MASs for concurrent task processing. 
In DEALS, each agent maintains local queues for unresolved tasks that arrive randomly. 
Based on its queue backlog and the differential in task completion success rates relative to its neighbors, each agent computes a score to determine whether to forward a task or process it locally. 
A key feature of DEALS is that task decomposition, allocation, and routing jointly account for both agent expertise and workload, without requiring additional LLM calls or a separately trained routing model. 
Moreover, each agent is equipped with multiple execution slots, enabling it to process independent tasks concurrently. 
Collectively, these decentralized interactions organize collaboration within individual tasks and coordinate execution across the workload without a central controller, while balancing answer quality and workload. 
Our contributions are summarized as follows:

\vspace{-.1in}
\begin{itemize}[leftmargin=*]
    \item \textbf{Self-Organizing Decentralized MAS Framework.}
We develop a decentralized MAS framework that enables self-organized collaboration and role specialization across concurrent tasks to achieve improvements in multiple dimensions, including accuracy, throughput, and latency. 
Specifically, each agent maintains a local queue for each task type and makes local routing decisions based on the queue state to direct task flows. Meanwhile, independent tasks can be executed concurrently both across agents and within individual agents. 
By jointly coordinating local routing, task decomposition, and concurrent execution, DEALS enables self-organized and self-evolving collaboration within tasks while efficiently allocating resources across the workload.

    \item \textbf{Lightweight Expertise-Aware Load-Balancing Routing.}
We design a decentralized expertise-aware backpressure-type routeing mechanism, which coordinates concurrent task execution while balancing answer quality and execution time.
Specifically, each router computes a routing score from backlog and task-specific success-rate differentials between the agent and its neighbors.
Each decision uses only local state and direct-neighbor information and requires neither additional LLM calls nor a separately trained routing model.

    \item \textbf{Empirical Validation.}
We evaluate DEALS across multiple benchmarks with homogeneous and heterogeneous agent pools.
The results show that the complete system achieves the highest average accuracy among the compared methods and $4.7\times$ to $22.7\times$ of the throughput of the baseline methods.
Further analysis shows that incorporating expertise information improves accuracy over backlog-only allocation and facilitates self-organization and self-evolution in heterogeneous MASs.
\end{itemize}

\section{Related Work}
\label{sec:related-work}

\noindent\textbf{1) LLM-based Multi-Agent Systems.}
MAS orchestrate LLM-based agents into structured workflows to solve tasks collaboratively \citep{guo2024multiagentsurvey,qi2026massurvey}.
Early systems structure collaboration through role-based conversations and predefined workflows \citep{li2023camel,wu2024autogen,hong2024metagpt}.
To improve answer quality, some works focus on debate or majority voting \citep{du2024multiagent,li2024moreagents}, which raise cost for each task.
Another line optimizes the collaboration structure itself.
For example, GPTSwarm optimizes agent computation graphs \citep{zhuge2024gptswarm}, while AFlow searches for effective agentic workflows \citep{zhang2025aflow}.
Furthermore, AgentPrune removes redundant communication to reduce token cost \citep{zhang2025agentprune}.
Beyond communication structure, DyLAN selects task-oriented teams and AgentVerse adjusts team composition during collaboration \citep{liu2024dylan,chen2024agentverse}.
These works considered solving one task at a time, whereas our method processes many independent tasks concurrently across agents.

\noindent\textbf{2) Centralized Routing for Task Allocation.}
Routing was first studied in the centralized setting, where a trained router assigns each query to one LLM from a candidate pool by balancing predicted answer quality against cost \citep{ding2024hybrid,ong2024routellm,feng2025graphrouter}.
MasRouter extends routing to the multi-agent setting for agent collaboration and role assignment \citep{yue2025masrouter}.
When independent tasks run concurrently, task allocation must consider both model ability and load balancing.
Toward this end, Chimera and RouteBalance combine model-quality estimates with runtime load \citep{ni2026chimera,da2026routebalance}.
However, both systems rely on separately trained predictors, which introduce additional training overhead.
All of these methods allocate work from a single decision point that needs global information of all agents, which limits scalability and requires each agent to expose its internal state.
Instead, our method enables agent to route tasks to neighbors by using {\em local state information and direct-neighbor communication}.

\noindent\textbf{3) Decentralized Coordination and Workload Allocation.}
In decentralized MAS, agents interact directly without a single orchestrator \citep{chen2025ioa,lu2024morphagent}.
Recent frameworks decide which agent solves a task, either from memory of trajectories \citep{yang2025agentnet,hao2026decentmem} or from a price that agents bid for the work \citep{bhatt2025coalesce,liu2026agentlance}.
Recent work draws on queueing and backpressure principles to schedule LLM inference and agent workloads \citep{dai2025throughput,zhang2026augmented}.
The resulting guarantees hold under specific assumptions that simplify real-world system dynamics.
Some methods select agents by capability or cost, and some consider queue backlogs and memory reuse when scheduling work.
Instead, our method combines success-rate differences with backlog differences for local routing.
Unlike AgentNet and AgentLance \citep{yang2025agentnet,liu2026agentlance}, our method requires no additional LLM calls for task allocation.

\par

\section{The Proposed DEALS Framework}
\label{sec:method}

\subsection{Problem Statement and System Model}

We consider a multi-agent system (MAS) whose topology is defined by a fixed directed graph $\mathcal{G}=(\mathcal{A},\mathcal{E})$, where $\mathcal{A}=\{a_1,\dots,a_n\}$ denotes the set of LLM agents, and $\mathcal{E}\subseteq\mathcal{A}\times\mathcal{A}$ denotes the connections between agents.
Tasks randomly arrive at one of the agents in the MAS system over time, and each task has a type $d\in\mathcal{D}$, where $\mathcal{D}$ denotes a finite task type set (e.g., coding, math, etc.).
Each agent $a_i\in\mathcal{A}$ contains three components: (i) a set of first-in, first-out (FIFO)  queueing buffers $\{Q_i^d\}_{d\in\mathcal{D}}$, which store tasks entering the system at agent $a_i$ and tasks forwarded by neighboring agents; (ii) a router,
which decides whether a task is executed locally or transferred to a neighbor;
and (iii) an executor,
which processes at most $C_i$ tasks concurrently and returns an answer.
The goal of the MAS is that each agent executes or decompose and routes tasks in a decentralized fashion, so that the system is self-organized, self-improve, and self-evolve to optimize performance in multiple dimensions (e.g., answer accuracy, throughput, and latency).

\begin{figure}[t!]
\centering
\includegraphics[width=\textwidth]{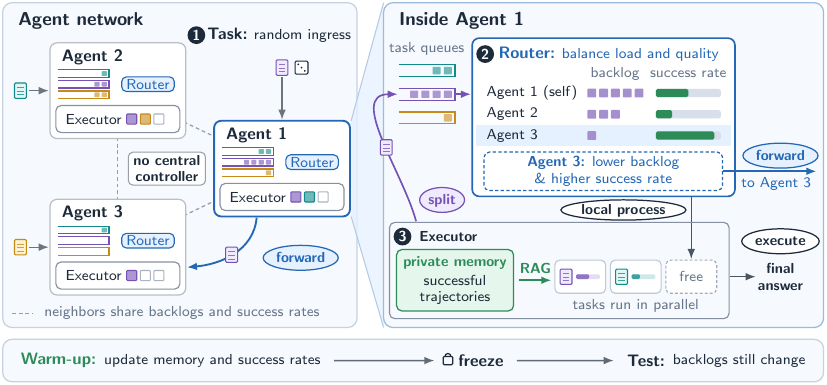}
\vspace{-.1in}
\caption[Overview of the DEALS framework.]{Overview of the DEALS framework.
\textbf{Left} shows collaboration among three agents without a central controller.
\textbf{Right} shows Agent~1's internal structure and explains its decision to forward the task to Agent~3.
\textbf{Bottom} shows how memories and success-rate estimates are updated during warm-up and frozen during testing, while backlogs continue to update.}
\label{fig:workflow}
\vspace{-.1in}
\end{figure}

Figure~\ref{fig:workflow} illustrates the task processing in DEALS, which is organized in three stages.
In Stage (i), a task of type-$d$ randomly arrives at an ingress agent $a$ in $\mathcal{A}$ following some stochastic arrival process %
and joins the type-$d$ queue of that agent.
Next, in Stage (ii), agent $a_i$ dequeues a task when service capacity is available and marks the task as ``in flight.''
Then, the router either forwards the task to the queue of a neighbor or assigns the task to local execution,
and a forwarded task repeats this stage at the receiving agent.
Lastly, in Stage (iii), a locally assigned task occupies one of the $C_i$ execution slots of agent $a_i$.
Each execution slot retrieves similar successful trajectories from the private memory of agent $a_i$ as demonstrations and then uses an execution policy (e.g., the ReAct execution policy~\citep{wei2022chain,yao2023react}).
Local execution subsequently returns a final answer or produces a continuation that re-enters the corresponding type-specific queue.
Once all queues are empty and no tasks remain in flight, the processing terminates.

\subsection{Decentralized Routing Design in DEALS}

We design a decentralized router that i) enables self-organization and self-evolution and ii) balances execution efficiency and answer quality.
Inspired by classical backpressure routing, our router uses backlog differentials to move tasks from more congested agents toward less congested neighbors \citep{tassiulas1992stability}.
However, backlog-only routing may favor an idle agent with low task-specific capability (e.g., in heterogeneous MAS).
Therefore, our router combines the backlog differential with the success-rate differential for each task type.
Different from existing works that process tasks sequentially based on expensive LLM nominations for routing (e.g., \citep{yang2025agentnet}), our method supports concurrent task processing both across agents and within each agent.
Each decision uses only the current state of the agent and the state reported by its direct neighbors.
In what follows, we present the key components of in the decentralized routing design in DEALS.

\noindent\textbf{1) Backlog.}
With a slight abuse of notation, we also use $Q_i^d(t)$ to denote the number of type-$d$ tasks stored in the queue $Q_i^d$ at time $t$.
Let $I_i^d(t)$ denote the number of dequeued type-$d$ tasks that are still being routed or executed by agent $a_i$.
The backlog $B_i^d(t)$ captures both waiting and active tasks and is defined as follows:
\begin{equation}
B_i^d(t)=Q_i^d(t)+I_i^d(t).
\label{eq:backlog}
\end{equation}
The backlog $B_i^d(t)$ remains unchanged when a task moves from the queue to local execution, and decreases only when the task is forwarded or completed.
Note that the classical backpressure formulation for routing in traditional data networks models does not include task execution at nodes.
However, local task execution in MAS requires non-negligible time.
Therefore, $B_i^d(t)$ augments the classical queueing backlog by explicitly considering tasks under execution at agent $a_i$.

\noindent\textbf{2) Task-Specific Quality.}
Backlog reflects workload but ignores each agent's problem-solving capability across different task types.
Therefore, we let each agent maintain a Beta-smoothed success rate for every task type.
Let $N_{i,d}^{+}$ and $N_{i,d}^{-}$ denote the numbers of successful and failed type-$d$ tasks executed by agent $a_i$.
The success rate of agent $a_i$ for task type $d$ can be estimated as:
\begin{equation}
q_i^d=
\frac{N_{i,d}^{+}+\alpha_q}
{N_{i,d}^{+}+N_{i,d}^{-}+\alpha_q+\beta_q}.
\label{eq:task-quality}
\end{equation}
The Beta prior can stabilize the estimate when few outcomes are available.
During the warm-up (training) split, every agent that executes or splits a task updates the corresponding success rate based on the verified outcome, while agents that only forward the task receive no update.
The resulting success rates are frozen before the test split.

\begin{figure}[t!]
\centering
\includegraphics[width=\textwidth]{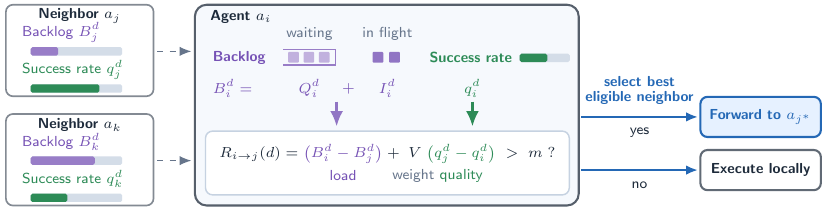}
\caption{Local routing in DEALS.
Agent $a_i$ combines the local state with neighbor advertisements to score eligible neighbors with backlog and success-rate differentials.
Agent $a_i$ forwards the task when the best score exceeds the margin $m$ and executes the task locally otherwise.}
\label{fig:router}
\vspace{-.2in}
\end{figure}

\noindent\textbf{3) Routing Score.}
To balance execution efficiency and answer quality, we combine the backlog differential $B_i^d-B_j^d$ and the {\em success-rate differential} $q_j^d-q_i^d$ into a routing score.
Given a nonnegative weight $V$, we define the routing score for forwarding a type-$d$ task from agent $a_i$ to neighbor $a_j$ as:
\begin{equation}
R_{i\rightarrow j}(d)
=B_i^d-B_j^d
+V\left(q_j^d-q_i^d\right).
\label{eq:routing-score}
\end{equation}

Intuitively, the routing score favors neighbors with lower backlog or a higher success rate for the type-$d$ task.
Specifically, $V=0$ removes the success-rate differential and considers only the backlog.
When $V\neq0$, a more capable neighbor can receive tasks despite a larger backlog until the backlog differential cancels the quality advantage.
Therefore, the router weighs congestion against capability rather than giving every agent the same number of tasks.

Figure~\ref{fig:router} shows how agent $a_i$ turns the local state and the neighbor advertisements into one routing decision.
For each type-$d$ task, agent $a_i$ evaluates all eligible direct neighbors and selects the neighbor with the highest routing score.
Agent $a_i$ forwards the task only when the maximum score exceeds margin $m$ and otherwise executes or splits the task locally.
The router excludes the previous agent and enforces hop limit $H$ to prevent immediate returns and unbounded forwarding.

\subsection{The Executor Design in DEALS}
\label{sec:executor}

\begingroup
Although our proposed DEALS is a general MAS framework, for fair performance comparisons, we adopt the ReAct-based retrieval-augmented executor that are widely used in the literature (e.g., \citep{yao2023react,yang2025agentnet}), while extending ReAct to support concurrent executions.
Specifically, DEALS equips each agent $a_i$ with $C_i$ execution slots for concurrently executing independent tasks both within and across agents.
At each step, the executor either completes the task or solves a subproblem and identifies the remaining work.
Figure~\ref{fig:workflow} illustrates the executor in an agent, which contains the following two key components:

\noindent\textbf{1) Memory-Guided Reasoning.}
Each agent $a_i$ maintains a private memory $M_i$ of successful trajectories.
These trajectories record the agent's local execution steps on warm-up tasks whose final answers are verified as correct.
The memory is frozen before testing and remains available for retrieval.
For each local step, the executor retrieves up to $k$ relevant successful trajectories from $M_i$ and uses them as demonstrations to guide its response.
The detailed calculation is deferred to Appendix~\ref{app:implementation}.

\noindent\textbf{2) Concurrent Execution.}
The executor supports two actions: \texttt{execute} produces a final answer, while \texttt{split} solves a subproblem and returns any unfinished work to the local queue.
Independent tasks run asynchronously, both in parallel across agents and concurrently within the $C_i$ execution slots of each agent $a_i$.
When a local step releases its slot, another waiting task can proceed without waiting for the other slots.
\par
\endgroup

\subsection{Overall Operations of DEALS}
\label{sec:overall}
\begingroup
\algrenewcommand\algorithmicrequire{\textbf{Input:}}
\algtext*{EndWhile}\algtext*{EndIf}\algtext*{EndFor}\algtext*{EndProcedure}
\begin{algorithm}[t]
\small
\caption{DEALS: Decentralized Expertise-Aware Load Serving at Agent $a_i$.}
\label{alg:deal}
\begin{algorithmic}[1]
\State \textbf{Input:} $\mathit{phase}\in\{\text{warm-up},\text{test}\}$, tasks $\mathcal T_i$ that enter the system at agent $a_i$, neighbors $\mathcal N(i)$
\State \textbf{Parameters:} capacity $C_i$, quality weight $V$, margin $m$, hop limit $H$, split limit $K$
\State \textbf{Initialization:} queue $Q_i^d\gets\varnothing$, in-flight count $I_i^d\gets0$, $\forall d\in\mathcal D$\label{line:alg-ingress} \Comment{Step~\Circled{1}: random ingress}
\If{$\mathit{phase}=\text{warm-up}$}
  \State memory $M_i\gets\varnothing$, success counts $N_{i,d}^{+}\gets0$, $N_{i,d}^{-}\gets0$, $\forall d\in\mathcal D$
\EndIf
\For{each type-$d$ task $x\in\mathcal T_i$}
  \State $\textsc{Enqueue}(Q_i^d,x)$, hop count $h(x)\gets0$, split count $k(x)\gets0$
  \State previous agent $\mathrm{prev}(x)\gets\varnothing$, executor set $S(x)\gets\varnothing$\label{line:alg-ingress-end}
\EndFor
\Procedure{Dispatch}{}\label{line:alg-dispatch}
  \While{$\sum_{d}I_i^d<C_i\land\exists d\colon Q_i^d\neq\varnothing$}
    \State $d\gets\arg\max_{d'}Q_i^{d'}$
    \State $x\gets\textsc{Dequeue}(Q_i^d)$, $I_i^d\gets I_i^d+1$
    \State \textbf{async} \Call{ProcessTask}{$x,d$}\label{line:alg-dispatch-end}
  \EndWhile
\EndProcedure
\Procedure{ProcessTask}{$x,d$}
  \State candidate set $\mathcal J\gets\{a_j\in\mathcal N(i)\mid a_j\neq\mathrm{prev}(x)\land h(x)<H\}$\label{line:alg-route} \Comment{Step~\Circled{2}: router}
  \State routing score $R_{i\to j}(d)\gets B_i^d-B_j^d+V(q_j^d-q_i^d)$, $\forall a_j\in\mathcal J$
  \If{$\mathcal J\neq\varnothing\land\max_{a_j\in\mathcal J}R_{i\to j}(d)>m$}
    \State best neighbor $a_{j^*}\gets\arg\max_{a_j\in\mathcal J}R_{i\to j}(d)$
    \State $\mathrm{prev}(x)\gets a_i$, $h(x)\gets h(x)+1$, $\textsc{Enqueue}(Q_{j^*}^d,x)$\label{line:alg-forward} \Comment{forward}
  \Else \Comment{Step~\Circled{3}: executor}
    \State $S(x)\gets S(x)\cup\{a_i\}$\label{line:alg-execute}
    \State allowed actions $\Omega(x)\gets\{\mathcal O_{\mathit{exec}}\}\cup\{\mathcal O_{\mathit{split}}\mid k(x)<K\}$
    \State $(\mathit{action},\mathit{output})\gets\textsc{ReAct}\bigl(x,\textsc{Retrieve}(M_i,x),\Omega(x)\bigr)$
    \If{$\mathit{action}=\mathcal O_{\mathit{split}}$}
      \State $x\gets\mathit{output}$, $k(x)\gets k(x)+1$, $\textsc{Enqueue}(Q_i^d,x)$
    \ElsIf{$\mathit{action}=\mathcal O_{\mathit{exec}}$}
      \State \Call{Finish}{$x,d,\mathit{output}$}
    \EndIf
  \EndIf
  \State $I_i^d\gets I_i^d-1$, \textsc{Dispatch}()\label{line:alg-release}
\EndProcedure
\Procedure{Finish}{$x,d,\mathit{answer}$}\label{line:alg-finish}
  \If{$\mathit{phase}=\text{warm-up}$}
    \State $\mathit{correct}\gets\textsc{Verify}(x,\mathit{answer})\in\{0,1\}$
    \For{each $a_j\in S(x)$}
      \State $N_{j,d}^{+}\gets N_{j,d}^{+}+\mathit{correct}$, $N_{j,d}^{-}\gets N_{j,d}^{-}+1-\mathit{correct}$
      \If{$\mathit{correct}=1$}
        \State $M_j\gets M_j\cup\{\textsc{Trajectory}(a_j,x)\}$
      \EndIf
    \EndFor
    \State \Return $\mathit{answer}$
  \ElsIf{$\mathit{phase}=\text{test}$}
    \State \Return $\mathit{answer}$\label{line:alg-finish-end}
  \EndIf
\EndProcedure
\end{algorithmic}
\end{algorithm}
\endgroup

Algorithm~\ref{alg:deal} summarizes DEALS at agent $a_i$, and every agent runs \textsc{Dispatch}, \textsc{ProcessTask} and \textsc{Finish} concurrently on local state as follows.

\noindent\textbf{1) Initialization.}
First, agent $a_i$ places each incoming task into the corresponding local task-type queue (Lines~\ref{line:alg-ingress}--\ref{line:alg-ingress-end}).
A type-$d$ task enters the system at agent $a_i$ with probability $p_i^d$, where $p^d=(p_1^d,\ldots,p_n^d)$ is an arbitrary probability distribution over the agents.

\noindent\textbf{2) Dispatch.}
When agent $a_i$ has at least one waiting task and an available execution slot, \textsc{Dispatch} admits a task from its longest typed queue (Lines~\ref{line:alg-dispatch}--\ref{line:alg-dispatch-end}).
Each admitted task is handled by a separate \textsc{ProcessTask} call, allowing up to $C_i$ tasks to progress concurrently within the agent.

\noindent\textbf{3) Task Processing.}
Next, \textsc{ProcessTask} scores the eligible neighbors with Eq.~\eqref{eq:routing-score} and forwards the task only when the best routing score exceeds the margin $m$ (Lines~\ref{line:alg-route}--\ref{line:alg-forward}).
If the task is retained locally, the executor uses ReAct with successful trajectories retrieved from $M_i$ to either return a final answer ($\mathcal O_{\mathit{exec}}$) or solve a subproblem and return the remaining work to the same typed queue ($\mathcal O_{\mathit{split}}$) (Lines~\ref{line:alg-execute}--\ref{line:alg-release}).
Different \textsc{ProcessTask} calls can run concurrently.
The hop limit $H$ and split limit $K$ bound the numbers of forwarding and split steps per task.

\noindent\textbf{4) Warm-up and Test.}
Finally, \textsc{Finish} verifies each warm-up answer and updates the success and failure counts of each agent in $S(x)$, the set of agents that executed part of task $x$ (Lines~\ref{line:alg-finish}--\ref{line:alg-finish-end}).
These agents add only their own successful trajectories to their private memories.
In contrast, the test phase only returns answers, and the memory and the success counts stay as learned in warm-up.
Backlogs continue to change during testing as tasks are queued and processed.
Thus, the router combines fixed learned quality estimates with changing workload information.

\section{Experimental Results}
\label{sec:experiments}

Our experiments examine both the performance of the complete system and the role of quality information in workload allocation.
First, we compare final-answer accuracy and report test throughput relative to baselines.
Second, we vary the quality weight to examine changes in accuracy and execution allocation.
Finally, we evaluate larger agent pools with different model compositions.

\subsection{Experimental setup}
\label{sec:experimental-setup}

\noindent\textbf{1) Tasks.}
    We evaluate on three benchmarks with task-type labels.
    For MATH and BBH, we reuse the released splits of \citet{yang2025agentnet}.
    We construct the MMLU-Pro split separately.
    \begin{itemize}[label={},leftmargin=0pt,itemindent=0pt,nosep]
        \item \textit{(i) MATH.} The split contains 700 training problems and 140 test problems across seven mathematical types \citep{hendrycks2021math}.
        \item \textit{(ii) BBH.} The split contains 627 training questions and 100 test questions drawn from 20 subtasks \citep{suzgun2022challenging}.
        \item \textit{(iii) MMLU-Pro.} We randomly sample 40 training questions and 20 test questions from each of five subjects: biology, computer science, history, law, and mathematics.
        This gives 200 training questions and 100 test questions in total \citep{wang2024mmlupro}.
    \end{itemize}
    For DEALS, we use Dirichlet-distributed ingress probabilities to model uneven task arrivals across agents.
    Appendix~\ref{app:implementation} gives the distribution parameters and sampling procedure.

\noindent\textbf{2) Models.}
    We use Qwen2.5-7B-Instruct, Mistral-7B-Instruct-v0.3, and Llama-3.1-8B-Instruct as agent backbones \citep{qwen2024qwen25,jiang2023mistral,grattafiori2024llama3}.
    For each backbone, we first evaluate a homogeneous group of three agents.
    We also evaluate a heterogeneous group with one agent from each model family.
    Next, to examine larger agent pools, we extend both settings to five and seven agents.
    The homogeneous pools H5 and H7 contain five and seven Qwen agents, respectively.
    In the heterogeneous setting, M5 combines three Qwen agents with one Mistral agent and one Llama agent.
    Building on M5, M7 adds two more Llama-based agents.

\noindent\textbf{3) Baselines.}
    We compare DEALS with single-agent and multi-agent frameworks as follows:
    \begin{itemize}[label={},leftmargin=0pt,itemindent=0pt,nosep]
        \item \textit{(i) Single Agent.} A single agent that uses ReAct and reasons step by step before acting \citep{wei2022chain,yao2023react}.
        \item \textit{(ii) AgentNet.} A decentralized multi-agent framework in which each agent uses the LLM to execute, split or forward a task based on agent abilities and retrieval-based memory \citep{yang2025agentnet}.
        \item \textit{(iii) GPTSwarm.} A multi-agent framework that models agents as optimizable computational graphs, learns the edges between agents and aggregates the answers by majority voting with benchmark-specific prompts \citep{zhuge2024gptswarm}.
        \item \textit{(iv) AgentPrune.} A multi-agent framework that prunes redundant messages from the communication graph between agents to reduce token cost \citep{zhang2025agentprune}.
    \end{itemize}

\Needspace{5\baselineskip}
\subsection{Accuracy and Throughput Analysis}

\begin{figure}[t]
\centering
\includegraphics[width=\textwidth]{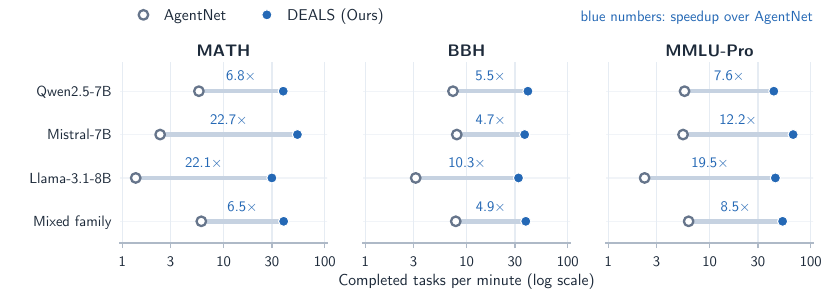}
\caption{Test throughput of sequential AgentNet and concurrent DEALS across three benchmarks and four backbone configurations.
Each row links AgentNet throughput (open circle) to DEALS throughput (filled circle), measured in completed tasks per test wall-clock minute on a log scale.
Blue numbers give the ratio of DEALS throughput to AgentNet throughput.
}
\label{fig:throughput}
\vspace{-.2in}
\end{figure}

As shown in Table~\ref{tab:main-results} and Figure~\ref{fig:throughput}, our method improves both accuracy and execution efficiency.
Since many existing methods lack public implementations or are costly to reproduce, we compare DEALS with the baselines listed in Table~\ref{tab:main-results}.

\noindent\textbf{1) Accuracy.}
\noindent
Table~\ref{tab:main-results} reports accuracy on MATH, BBH, and MMLU-Pro across four agent configurations.
DEALS achieves the highest average accuracy in all four configurations and ranks first in nine of the twelve individual evaluations.
It also outperforms AgentNet on all three benchmarks in every configuration.
The largest average improvement over the strongest baseline occurs in the heterogeneous setting, where DEALS reaches $66.52\%$ compared with $57.76\%$ for AgentNet.

\begin{table}[t]
\caption{Accuracy (\%) across backbones and datasets.
Bold indicates the best result, and underlining indicates the second-best result.
The quality weight is set to $V=20$ for homogeneous configurations and $V=80$ for the mixed-family configuration.}
\label{tab:main-results}
\centering
\scriptsize
\setlength{\tabcolsep}{5.0pt}
\renewcommand{\arraystretch}{0.9}
\resizebox{0.8\linewidth}{!}{%
\begin{tabular}{llcccc}
\toprule
Backbone & System & MATH & BBH & MMLU-Pro & Average \\
\midrule
\multirow{5}{*}{Qwen2.5-7B}
& Single Agent & 66.43 & 62.00 & \underline{52.00} & 60.14 \\
& AgentNet & \underline{68.57} & 71.00 & 50.00 & \underline{63.19} \\
& GPTSwarm & 65.00 & \textbf{75.00} & 21.00 & 53.67 \\
& AgentPrune & 65.00 & 68.00 & 34.00 & 55.67 \\
\rowcolor{blue!8}
& DEALS (Ours) & \textbf{71.43} & \underline{74.00} & \textbf{53.00} & \textbf{66.14} \\
\midrule
\multirow{5}{*}{Mistral-7B}
& Single Agent & 6.43 & 43.00 & 34.00 & 27.81 \\
& AgentNet & 7.86 & \underline{50.00} & 30.00 & \underline{29.29} \\
& GPTSwarm & \textbf{13.57} & 30.00 & \underline{35.00} & 26.19 \\
& AgentPrune & 6.43 & 14.00 & 19.00 & 13.14 \\
\rowcolor{blue!8}
& DEALS (Ours) & \underline{12.86} & \textbf{51.00} & \textbf{38.00} & \textbf{33.95} \\
\midrule
\multirow{5}{*}{Llama-3.1-8B}
& Single Agent & \underline{52.86} & \underline{53.00} & 40.00 & 48.62 \\
& AgentNet & 31.43 & \underline{53.00} & 27.00 & 37.14 \\
& GPTSwarm & \textbf{57.14} & 50.00 & \underline{43.00} & \underline{50.05} \\
& AgentPrune & 33.57 & 49.00 & 35.00 & 39.19 \\
\rowcolor{blue!8}
& DEALS (Ours) & 51.43 & \textbf{60.00} & \textbf{48.00} & \textbf{53.14} \\
\midrule
\multirow{4}{*}{Mixed Family} & AgentNet & \underline{64.29} & 57.00 & \underline{52.00} & \underline{57.76} \\
& GPTSwarm & 63.57 & 61.00 & 38.00 & 54.19 \\
& AgentPrune & 50.71 & \underline{65.00} & 37.00 & 50.90 \\
\rowcolor{blue!8}
& DEALS (Ours) & \textbf{73.57} & \textbf{73.00} & \textbf{53.00} & \textbf{66.52} \\
\bottomrule
\end{tabular}
}
\vspace{-0.35\baselineskip}
\end{table}

\noindent\textbf{2) Throughput.}
Figure~\ref{fig:throughput} reports throughput measured as completed tasks per minute.
Across all settings, the throughput of our method is $4.7\times$ to $22.7\times$ that of AgentNet.
This gain comes from {\em concurrent} task execution and decentralized workload allocation.
Therefore, multiple agents work simultaneously and each agent can execute up to $C_i$ tasks concurrently.
In contrast, AgentNet processes one task at a time which limits throughput.

\subsection{Guiding Self-Organization through Local Routing}
\label{sec:quality-sensitivity}

We vary the quality weight $V$ to examine how local routing decisions shape task allocation across the MAS.
This experiment also examines whether changing the routing preference can guide self-organization without prescribing individual task assignments and agent role specialization.

\begin{figure}[t!]
\centering
\begin{subfigure}{\textwidth}
\centering
\includegraphics[width=\textwidth]{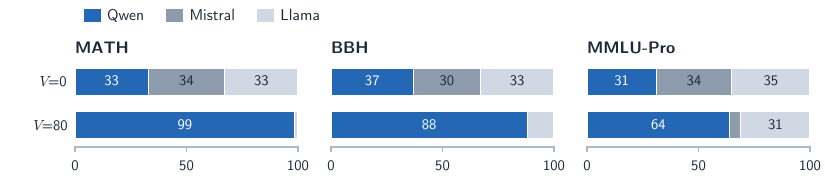}
\caption{Final-answer percentage by agent in the heterogeneous pool.}
\label{fig:quality-weight-allocation}
\end{subfigure}

\vspace{0.5\baselineskip}
\begin{subfigure}{\textwidth}
\centering
\includegraphics[width=\textwidth]{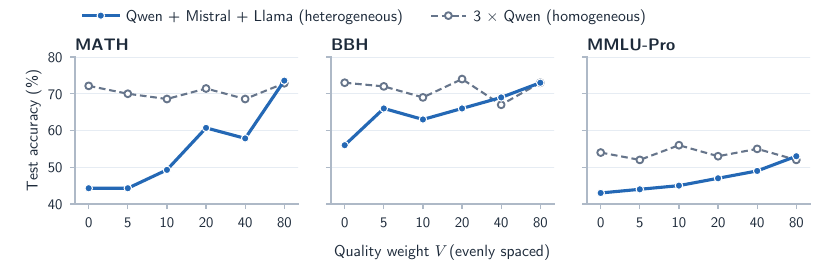}
\caption{Test accuracy versus the quality weight $V$.}
\label{fig:quality-weight-accuracy}
\end{subfigure}
\caption{Effect of the quality weight $V$ in three-agent systems.
(a) Final-answer percentage by agent in the heterogeneous pool for $V=0$ and $V=80$.
(b) Test accuracy of the heterogeneous and the homogeneous pools at different quality weights $V$.}
\label{fig:quality-weight-analysis}
\vspace{-.2in}
\end{figure}

\noindent\textbf{1) Self-Organized Task Allocation.}
Figure~\ref{fig:quality-weight-analysis}(a) shows the percentage of final answers produced by each agent.
At $V=0$, the router uses backlog alone without considering task-solving capability, and final answers are distributed approximately evenly among the three agents.
At $V=80$, the router shifts final task executions toward the stronger Qwen agent.
For example, Qwen produces approximately $99\%$ of the final answers on MATH and $64\%$ on MMLU-Pro.
These allocation patterns arise from local routing decisions rather than predefined  assignment policy.
It also illustrate how agents organize task completion through local interactions.

\noindent\textbf{2) Controlling the Expertise-Load Balance.}
The weight $V$ allows users to adjust the relative importance of agent expertise and workload in local routing.
Figure~\ref{fig:quality-weight-analysis}(b) examines how this preference affects answer accuracy.
In the homogeneous setting, accuracy varies less with $V$ because agents share the same backbone and have more similar task-specific capabilities.
However, in the heterogeneous pool, $V=0$ gives the lowest accuracy on all three datasets.
Increasing $V$ from $0$ to $80$ then improves accuracy on every dataset, for example from $44.29\%$ to $73.57\%$ on MATH.
These results show that task allocation benefits from quality information instead of relying on backlog alone.

\subsection{Expertise-aware routing at larger pool sizes}
\label{sec:agent-count}

Following the experimental setup in Section~\ref{sec:experimental-setup}, we evaluate DEALS on MATH with five and seven agents, using Qwen-only homogeneous pools (H5 and H7) and heterogeneous pools that combine Qwen with Mistral and Llama (M5 and M7).
Table~\ref{tab:agent-count-accuracy} shows that expertise-aware routing improves accuracy in both heterogeneous pools.
Increasing $V$ from $0$ to $20$ and then to $80$ yields substantial accuracy gains.
In contrast, homogeneous accuracy remains  stable across different values of $V$.

\begin{wraptable}[13]{r}{0.49\textwidth}
\vspace{-1.1\baselineskip}
\caption{MATH test accuracy (\%) with five and seven agents.
Bold marks the highest accuracy within each group, including ties.}
\label{tab:agent-count-accuracy}
\centering
\small
\setlength{\tabcolsep}{3pt}
\begin{tabular}{llrrr}
\toprule
Setting & Group & $V=0$ & $V=20$ & $V=80$ \\
\midrule
\multirow{2}{*}{Homogeneous}
& H5 & \textbf{68.57} & \textbf{68.57} & \textbf{68.57} \\
& H7 & 70.71 & 70.71 & \textbf{72.14} \\
\midrule
\multirow{2}{*}{Heterogeneous}
& M5 & 47.86 & 64.29 & \textbf{72.86} \\
& M7 & 40.00 & 60.71 & \textbf{67.86} \\
\bottomrule
\end{tabular}

\end{wraptable}
Both M5 and M7 contain Mistral and Llama agents that are weaker than Qwen on MATH.
With backlog-only routing, M5 and M7 reach $47.86\%$ and $40.00\%$, which falls below the $66.43\%$ of the strongest single agent in Table~\ref{tab:main-results}.
With $V=80$, the same two pools reach $72.86\%$ and $67.86\%$ and pass that single agent.
Accuracy drops from M5 to M7 because M7 adds two weaker Llama agents, yet M7 still passes that single agent.
The homogeneous pools change little, because every agent shares the same backbone and leaves little quality differential for the router to exploit.

\FloatBarrier
\vspace{-0.35\baselineskip}
\section{Conclusion}
\label{sec:conclusion}

In this paper, we developed DEALS for decentralized coordination for serving concurrent task workloads in multi-agent systems.
The key principle is that available execution capacity should be considered together with task-solving capability.
DEALS implements this principle through local backlog and success-rate differentials while independent tasks proceed asynchronously.
Our experiments showed that incorporating quality information is particularly useful in heterogeneous pools.
Furthermore, our framework achieved competitive accuracy and higher workload throughput than state-of-the-art baselines.

\bibliographystyle{iclr2027_conference}
\bibliography{references}

@article{yang2025agentnet,
  title={Agentnet: Decentralized evolutionary coordination for llm-based multi-agent systems},
  author={Yang, Yingxuan and Chai, Huacan and Shao, Shuai and Song, Yuanyi and Qi, Siyuan and Rui, Renting and Zhang, Weinan},
  journal={Advances in Neural Information Processing Systems},
  volume={38},
  pages={107309--107336},
  year={2026}
}

@inproceedings{tassiulas1992stability,
  title={Stability properties of constrained queueing systems and scheduling policies for maximum throughput in multihop radio networks},
  author={Tassiulas, Leandros and Ephremides, Anthony},
  booktitle={29th IEEE Conference on Decision and Control},
  pages={2130--2132},
  year={1990},
  organization={IEEE}
}

@article{hendrycks2021math,
  title={Measuring mathematical problem solving with the math dataset},
  author={Hendrycks, Dan and Burns, Collin and Kadavath, Saurav and Arora, Akul and Basart, Steven and Tang, Eric and Song, Dawn and Steinhardt, Jacob},
  journal={arXiv preprint arXiv:2103.03874},
  year={2021}
}

@inproceedings{suzgun2022challenging,
  title={Challenging big-bench tasks and whether chain-of-thought can solve them},
  author={Suzgun, Mirac and Scales, Nathan and Sch{\"a}rli, Nathanael and Gehrmann, Sebastian and Tay, Yi and Chung, Hyung Won and Chowdhery, Aakanksha and Le, Quoc and Chi, Ed H and Zhou, Denny and others},
  booktitle={Findings of the Association for Computational Linguistics: ACL 2023},
  pages={13003--13051},
  year={2023}
}

@article{wang2024mmlupro,
  title={Mmlu-pro: A more robust and challenging multi-task language understanding benchmark},
  author={Wang, Yubo and Ma, Xueguang and Zhang, Ge and Ni, Yuansheng and Chandra, Abhranil and Guo, Shiguang and Ren, Weiming and Arulraj, Aaran and He, Xuan and Jiang, Ziyan and others},
  journal={Advances in Neural Information Processing Systems},
  volume={37},
  pages={95266--95290},
  year={2024}
}

@article{qwen2024qwen25,
  title={Qwen2. 5-math technical report: Toward mathematical expert model via self-improvement},
  author={Yang, An and Zhang, Beichen and Hui, Binyuan and Gao, Bofei and Yu, Bowen and Li, Chengpeng and Liu, Dayiheng and Tu, Jianhong and Zhou, Jingren and Lin, Junyang and others},
  journal={arXiv preprint arXiv:2409.12122},
  year={2024}
}

@inproceedings{jiang2023mistral,
  title={Comprehensive examination of instruction-based language models: A comparative analysis of mistral-7b and llama-2-7b},
  author={Thakkar, Hiren and Manimaran, A},
  booktitle={2023 International Conference on Emerging Research in Computational Science (ICERCS)},
  pages={1--6},
  year={2023},
  organization={IEEE}
}

@article{grattafiori2024llama3,
  title={The llama 3 herd of models},
  author={Grattafiori, Aaron and Dubey, Abhimanyu and Jauhri, Abhinav and Pandey, Abhinav and Kadian, Abhishek and Al-Dahle, Ahmad and Letman, Aiesha and Mathur, Akhil and Schelten, Alan and Vaughan, Alex and others},
  journal={arXiv preprint arXiv:2407.21783},
  year={2024}
}

@article{brown2020language,
  title={Language models are few-shot learners},
  author={Brown, Tom and Mann, Benjamin and Ryder, Nick and Subbiah, Melanie and Kaplan, Jared D and Dhariwal, Prafulla and Neelakantan, Arvind and Shyam, Pranav and Sastry, Girish and Askell, Amanda and others},
  journal={Advances in neural information processing systems},
  volume={33},
  pages={1877--1901},
  year={2020}
}

@article{liu2024lost,
  title={Lost in the middle: How language models use long contexts},
  author={Liu, Nelson F and Lin, Kevin and Hewitt, John and Paranjape, Ashwin and Bevilacqua, Michele and Petroni, Fabio and Liang, Percy},
  journal={Transactions of the association for computational linguistics},
  volume={12},
  pages={157--173},
  year={2024}
}

@inproceedings{liu2024agentbench,
  title={Agentbench: Evaluating llms as agents},
  author={Liu, Xiao and Yu, Hao and Zhang, Hanchen and Xu, Yifan and Lei, Xuanyu and Lai, Hanyu and Gu, Yu and Ding, Hangliang and Men, Kaiwen and Yang, Kejuan and others},
  booktitle={International Conference on Learning Representations},
  volume={2024},
  pages={52989--53046},
  year={2024}
}

@article{wei2022chain,
  title={Chain-of-thought prompting elicits reasoning in large language models},
  author={Wei, Jason and Wang, Xuezhi and Schuurmans, Dale and Bosma, Maarten and Xia, Fei and Chi, Ed and Le, Quoc V and Zhou, Denny and others},
  journal={Advances in neural information processing systems},
  volume={35},
  pages={24824--24837},
  year={2022}
}

@article{yao2023react,
  title={React: Synergizing reasoning and acting in language models},
  author={Yao, Shunyu and Zhao, Jeffrey and Yu, Dian and Du, Nan and Shafran, Izhak and Narasimhan, Karthik and Cao, Yuan},
  journal={arXiv preprint arXiv:2210.03629},
  year={2022}
}

@inproceedings{xiao2023cpack,
  title={C-pack: Packed resources for general chinese embeddings},
  author={Xiao, Shitao and Liu, Zheng and Zhang, Peitian and Muennighoff, Niklas and Lian, Defu and Nie, Jian-Yun},
  booktitle={Proceedings of the 47th international ACM SIGIR conference on research and development in information retrieval},
  pages={641--649},
  year={2024}
}

@article{dai2025throughput,
  title={Throughput-optimal scheduling algorithms for llm inference and ai agents},
  author={Dai, JG and Deng, Tianze and Li, Yueying and Peng, Tianyi},
  journal={arXiv preprint arXiv:2504.07347},
  year={2025}
}

@article{luo2025autellix,
  title={Autellix: An efficient serving engine for llm agents as general programs},
  author={Luo, Michael and Shi, Xiaoxiang and Cai, Colin and Zhang, Tianjun and Wong, Justin and Wang, Yichuan and Wang, Chi and Huang, Yanping and Chen, Zhifeng and Gonzalez, Joseph E and others},
  journal={arXiv preprint arXiv:2502.13965},
  year={2025}
}

@inproceedings{lin2024parrot,
  title={Parrot: Efficient serving of $\{$LLM-based$\}$ applications with semantic variable},
  author={Lin, Chaofan and Han, Zhenhua and Zhang, Chengruidong and Yang, Yuqing and Yang, Fan and Chen, Chen and Qiu, Lili},
  booktitle={18th USENIX Symposium on Operating Systems Design and Implementation (OSDI 24)},
  pages={929--945},
  year={2024}
}

@article{li2023camel,
  title={Camel: Communicative agents for" mind" exploration of large language model society},
  author={Li, Guohao and Hammoud, Hasan and Itani, Hani and Khizbullin, Dmitrii and Ghanem, Bernard},
  journal={Advances in neural information processing systems},
  volume={36},
  pages={51991--52008},
  year={2023}
}

@article{wu2024autogen,
  title={Autogen: Enabling next-gen llm applications via multi-agent conversation},
  author={Wu, Qingyun and Bansal, Gagan and Zhang, Jieyu and Wu, Yiran and Li, Beibin and Zhu, Erkang and Jiang, Li and Zhang, Xiaoyun and Zhang, Shaokun and Liu, Jiale and others},
  journal={arXiv preprint arXiv:2308.08155},
  year={2023}
}

@inproceedings{hong2024metagpt,
  title={MetaGPT: Meta programming for a multi-agent collaborative framework},
  author={Hong, Sirui and Zhuge, Mingchen and Chen, Jonathan and Zheng, Xiawu and Cheng, Yuheng and Wang, Jinlin and Zhang, Ceyao and Yau, Steven and Lin, Zijuan and Zhou, Liyang and others},
  booktitle={International Conference on Learning Representations},
  volume={2024},
  pages={23247--23275},
  year={2024}
}

@inproceedings{yue2025masrouter,
  title={Masrouter: Learning to route llms for multi-agent systems},
  author={Yue, Yanwei and Zhang, Guibin and Liu, Boyang and Wan, Guancheng and Wang, Kun and Cheng, Dawei and Qi, Yiyan},
  booktitle={Proceedings of the 63rd Annual Meeting of the Association for Computational Linguistics (Volume 1: Long Papers)},
  pages={15549--15572},
  year={2025}
}

@article{ni2026chimera,
  title={Chimera: Latency-and Performance-Aware Multi-agent Serving for Heterogeneous LLMs},
  author={Ni, Kangqi and Hua, Wenyue and Shi, Xiaoxiang and Guo, Jiang and Chang, Shiyu and Chen, Tianlong},
  journal={arXiv preprint arXiv:2603.22206},
  year={2026}
}

@article{zhang2026augmented,
  title={Augmented Backpressure for Decentralized Management of Agentic Networks},
  author={Zhang, Zuyuan and Tang, Sizhe and Lan, Tian},
  journal={arXiv preprint arXiv:2608.00914},
  year={2026}
}

@article{zhuge2024gptswarm,
  title={Language agents as optimizable graphs},
  author={Zhuge, Mingchen and Wang, Wenyi and Kirsch, Louis and Faccio, Francesco and Khizbullin, Dmitrii and Schmidhuber, J{\"u}rgen},
  journal={arXiv preprint arXiv:2402.16823},
  year={2024}
}

@inproceedings{zhang2025aflow,
  title={Aflow: Automating agentic workflow generation},
  author={Zhang, Jiayi and Xiang, Jinyu and Yu, Zhaoyang and Teng, Fengwei and Chen, Xionghui and Chen, Jiaqi and Zhuge, Mingchen and Cheng, Xin and Hong, Sirui and Wang, Jinlin and others},
  booktitle={International Conference on Learning Representations},
  volume={2025},
  pages={34040--34077},
  year={2025}
}

@inproceedings{zhang2025agentprune,
  title={Cut the crap: An economical communication pipeline for llm-based multi-agent systems},
  author={Zhang, Guibin and Yue, Yanwei and Li, Zhixun and Yun, Sukwon and Wan, Guancheng and Wang, Kun and Cheng, Dawei and Yu, Jeffrey and Chen, Tianlong},
  booktitle={International Conference on Learning Representations},
  volume={2025},
  pages={75389--75428},
  year={2025}
}

@article{liu2024dylan,
  title={A dynamic LLM-powered agent network for task-oriented agent collaboration},
  author={Liu, Zijun and Zhang, Yanzhe and Li, Peng and Liu, Yang and Yang, Diyi},
  journal={arXiv preprint arXiv:2310.02170},
  year={2023}
}

@article{du2024multiagent,
  title={Improving factuality and reasoning in language models through multiagent debate},
  author={Du, Yilun and Li, Shuang and Torralba, Antonio and Tenenbaum, Joshua B and Mordatch, Igor},
  journal={arXiv preprint arXiv:2305.14325},
  year={2023}
}

@inproceedings{kwon2023vllm,
  title={Efficient memory management for large language model serving with pagedattention},
  author={Kwon, Woosuk and Li, Zhuohan and Zhuang, Siyuan and Sheng, Ying and Zheng, Lianmin and Yu, Cody Hao and Gonzalez, Joseph and Zhang, Hao and Stoica, Ion},
  booktitle={Proceedings of the 29th symposium on operating systems principles},
  pages={611--626},
  year={2023}
}

@inproceedings{ding2024hybrid,
  title={Hybrid llm: Cost-efficient and quality-aware query routing},
  author={Ding, Dujian and Mallick, Ankur and Wang, Chi and Sim, Robert and Mukherjee, Subhabrata and R{\"u}hle, Victor and Lakshmanan, Laks and Awadallah, Ahmed H},
  booktitle={International Conference on Learning Representations},
  volume={2024},
  pages={41348--41366},
  year={2024}
}

@inproceedings{ong2024routellm,
  title={Routellm: Learning to route llms from preference data},
  author={Ong, Isaac and Almahairi, Amjad and Wu, Vincent and Chiang, Wei-Lin and Wu, Tianhao and Gonzalez, Joseph E and Kadous, Mohammed and Stoica, Ion},
  booktitle={International Conference on Learning Representations},
  volume={2025},
  pages={34433--34448},
  year={2025}
}

@inproceedings{feng2025graphrouter,
  title={Graphrouter: A graph-based router for llm selections},
  author={Feng, Tao and Shen, Yanzhen and You, Jiaxuan},
  booktitle={International Conference on Learning Representations},
  volume={2025},
  pages={26186--26203},
  year={2025}
}

@inproceedings{chen2025ioa,
  title={Internet of agents: Weaving a web of heterogeneous agents for collaborative intelligence},
  author={Chen, Weize and You, Ziming and Li, Ran and Qian, Chen and Zhao, Chenyang and Yang, Cheng and Xie, Ruobing and Liu, Zhiyuan and Sun, Maosong and others},
  booktitle={International Conference on Learning Representations},
  volume={2025},
  pages={36374--36411},
  year={2025}
}

@article{lu2024morphagent,
  title={Morphagent: Empowering agents through self-evolving profiles and decentralized collaboration},
  author={Lu, Siyuan and Shao, Jiaqi and Luo, Bing and Lin, Tao},
  journal={arXiv preprint arXiv:2410.15048},
  year={2024}
}

@article{hao2026decentmem,
  title={Self-Evolving Multi-Agent Systems via Decentralized Memory},
  author={Hao, Guangya and Long, Yunbo and Zhao, Zhuokai},
  journal={arXiv preprint arXiv:2605.22721},
  year={2026}
}

@article{da2026routebalance,
  title={RouteBalance: Fused Model Routing and Load Balancing for Heterogeneous LLM Serving},
  author={Da, Wei and Kalyvianaki, Evangelia},
  journal={arXiv preprint arXiv:2606.17949},
  year={2026}
}

@article{liu2026agentlance,
  title={Markets, Not Planners: Decentralized Orchestration of LLM Agents with Private Information},
  author={Liu, Xiao and Li, Haoyang and Li, Songwei and Fang, Hongbo and Xu, Fengli and Shi, Feng and Evans, James},
  journal={arXiv preprint arXiv:2608.23867},
  year={2026}
}

@article{guo2024multiagentsurvey,
  title={Large language model based multi-agents: A survey of progress and challenges},
  author={Guo, Taicheng and Chen, Xiuying and Wang, Yaqi and Chang, Ruidi and Pei, Shichao and Chawla, Nitesh V and Wiest, Olaf and Zhang, Xiangliang},
  journal={arXiv preprint arXiv:2402.01680},
  year={2024}
}

@inproceedings{chen2024agentverse,
  title={Agentverse: Facilitating multi-agent collaboration and exploring emergent behaviors},
  author={Chen, Weize and Su, Yusheng and Zuo, Jingwei and Yang, Cheng and Yuan, Chenfei and Chan, Chi-Min and Yu, Heyang and Lu, Yaxi and Hung, Yi-Hsin and Qian, Chen and others},
  booktitle={International Conference on Learning Representations},
  volume={2024},
  pages={20094--20136},
  year={2024}
}

@article{qi2026massurvey,
  title={Beyond Individual Intelligence: Surveying Collaboration, Failure Attribution, and Self-Evolution in LLM-based Multi-Agent Systems},
  author={Qi, Shihao and Ma, Jie and Xing, Rui and Guo, Wei and Huang, Xiao and Gao, Zhitao and Deng, Jianhao and Liu, Jun and Zhang, Lingling and Wei, Bifan and others},
  journal={arXiv preprint arXiv:2605.14892},
  year={2026}
}

@article{li2024moreagents,
  title={More agents is all you need},
  author={Li, Junyou and Zhang, Qin and Yu, Yangbin and Fu, Qiang and Ye, Deheng},
  journal={arXiv preprint arXiv:2402.05120},
  year={2024}
}

@inproceedings{bhatt2025coalesce,
  title={Coalesce: Economic and security dynamics of skill-based task outsourcing among team of autonomous llm agents},
  author={Bhatt, Manish and Del Rosario, Ronald F and Narajala, Vineeth Sai and Habler, Idan},
  booktitle={2025 Cyber Awareness and Research Symposium (CARS)},
  pages={1--9},
  year={2025},
  organization={IEEE}
}

@article{zhang2025landscape,
  title={The landscape of agentic reinforcement learning for llms: A survey},
  author={Zhang, Guibin and Geng, Hejia and Yu, Xiaohang and Yin, Zhenfei and Zhang, Zaibin and Tan, Zelin and Zhou, Heng and Li, Zhongzhi and Xue, Xiangyuan and Li, Yijiang and others},
  journal={arXiv preprint arXiv:2509.02547},
  year={2025}
}

\appendix

\newpage
\section{Implementation details}
\label{app:implementation}

\begingroup
\color{black}
This appendix describes how DEALS runs in our experiments, and Table~\ref{tab:implementation} lists the parameter settings.
\begin{table}[H]
\caption{Implementation parameter settings for all experiments.}
\label{tab:implementation}
\centering
\small
\setlength{\tabcolsep}{4pt}
\begin{tabular}{@{}p{0.46\linewidth}p{0.42\linewidth}@{}}
\toprule
Parameter & Setting \\
\midrule
Number of agents & $3$, $5$, $7$ \\
Graph topology & Complete graph \\
Dirichlet concentration $\alpha$ & $0.7$ \\
Execution capacity per agent $C_i$ & $9$ \\
Hop limit $H$ & $3$ \\
Split limit $K$ & $3$ \\
Forwarding margin $m$ & $1$ \\
Beta prior $(\alpha_q,\beta_q)$ & $(1,1)$ \\
Quality weight $V$ & $20$ (homogeneous), $80$ (mixed family) \\
Temperature & $0$ \\
Request seed & $1$ \\
Hardware & $8\times$ A100 80GB (five and seven agents), $4\times$ H100 NVL or $4\times$ H200 NVL (three agents) \\
\bottomrule
\end{tabular}
\end{table}
\noindent\textbf{Serving.}
Each agent serves the backbone model through a private vLLM replica on a dedicated GPU \citep{kwon2023vllm}.
A separate GPU hosts the BGE-large retrieval encoder \citep{xiao2023cpack}.
All model calls use temperature $0$, request seed $1$, and at most $2048$ generated tokens.

\noindent\textbf{Concurrency.}
Each agent processes tasks with $C_i=9$ worker slots, and a task holds one slot while agent $a_i$ routes or executes the task.
Whenever a slot becomes free, agent $a_i$ admits the next task from the longest typed queue and breaks ties by the order of task types.

\noindent\textbf{Routing.}
The router evaluates Eq.~\eqref{eq:routing-score} for every direct neighbor except the previous agent of the task.
The task moves to the best neighbor only when the best score is strictly larger than $m=1$, and ties go to the neighbor with the smallest index.
The hop limit $H=3$ counts every forward of a task, including forwards after a split.
Once a task reaches the hop limit, the task runs at the agent that currently holds the task.

\begingroup
\color{black}
\noindent\textbf{Trajectory retrieval.}
Let $e(\cdot)$ denote the BGE-large text embedding \citep{xiao2023cpack}, and let $t(\tau)$ denote the recorded execution time of a successful trajectory $\tau$ in minutes.
For a query text $z$, we define the relevance of trajectory $\tau$ as
\begin{equation}
s(z,\tau)=\cos\bigl(e(z),e(\tau)\bigr)\left(1+\frac{0.2}{1+t(\tau)}\right),
\label{eq:relevance}
\end{equation}
where $e(\tau)$ embeds the stored task context or thought.
The second factor favors trajectories with shorter execution times.
The executor retrieves relevant successful trajectories as follows:
\begin{equation}
\textsc{Retrieve}(M_i,z)=\operatorname{top}_k\bigl\{\tau\in M_i\bigm| s(z,\tau)\ge\theta\bigr\},
\label{eq:retrieve}
\end{equation}
where $\operatorname{top}_k$ selects up to $k$ trajectories with the highest relevance scores of at least $\theta$.

Equation~\eqref{eq:relevance} combines cosine similarity with a time-based factor whose coefficient is fixed at $0.2$.
Here, $z$ is the task context or a preliminary thought, and $e(\tau)$ embeds the corresponding stored context or thought.
The recorded time $t(\tau)$ covers local trajectory execution and excludes action selection, preliminary-thought generation and completion checks.
In Eq.~\eqref{eq:retrieve}, $M_i$ is the agent's private memory of successful trajectories.
We set $\theta=0.7$ and $k=3$, so retrieval returns at most three trajectories that meet the relevance threshold.
If fewer than three trajectories qualify, all qualifying trajectories are returned.
The retrievable execution pool of each agent holds at most $40$ successful trajectories.
\par
\endgroup

\noindent\textbf{Warm-up and test.}
In our experiments, ingress probabilities follow $p^d\sim\mathrm{Dirichlet}(\alpha\mathbf{1}_n)$ with $\alpha=0.7$.
We draw one vector per task type in each phase and keep it fixed within that phase.
Warm-up and testing use different random seeds for ingress.
During warm-up, each finished task is verified immediately, so later routing decisions already use the updated success rates.
During test, the memories and the success rates stay fixed, and answers are scored only after all test tasks finish.

\par
\endgroup

\section{Limitations}
\label{app:limitations}

Our evaluation covers complete graphs with three, five, and seven agents across three LLM families.
Future work can study other communication topologies such as chain, ring, and sparse graphs.
Furthermore, future work can evaluate more challenging benchmarks and include a broader range of LLM families.

\section{Extended results}
\label{app:extended}

\subsection{Quality-weight sensitivity and final-task allocation}
\label{app:quality-weight-data}

Tables~\ref{tab:v-sensitivity} and~\ref{tab:heterogeneous-allocation} preserve the exact numerical values used in Figure~\ref{fig:quality-weight-analysis}.

\begin{table}[htbp]
\caption{Accuracy (\%) sensitivity to the quality weight $V$.
Spread is the difference between the highest and lowest accuracy across the reported values of $V$.
Bold indicates the best result within each row.}
\label{tab:v-sensitivity}
\centering
\normalsize
\setlength{\tabcolsep}{3.0pt}
\begin{tabular}{llccccccc}
\toprule
Agent pool & Dataset & $V{=}0$ & $V{=}5$ & $V{=}10$ & $V{=}20$ & $V{=}40$ & $V{=}80$ & Spread \\
\midrule
\multirow{3}{*}{\shortstack[l]{Homogeneous\\Qwen2.5-7B $\times 3$}}
& MATH & \cellcolor{blue!9}72.14 & \cellcolor{blue!3}70.00 & \cellcolor{blue!0}68.57 & \cellcolor{blue!6}71.43 & \cellcolor{blue!0}68.57 & \cellcolor{blue!12}\textbf{72.86} & 4.29 \\
& BBH & \cellcolor{blue!9}73.00 & \cellcolor{blue!6}72.00 & \cellcolor{blue!3}69.00 & \cellcolor{blue!12}\textbf{74.00} & \cellcolor{blue!0}67.00 & \cellcolor{blue!9}73.00 & 7.00 \\
& MMLU-Pro & \cellcolor{blue!6}54.00 & \cellcolor{blue!0}52.00 & \cellcolor{blue!12}\textbf{56.00} & \cellcolor{blue!3}53.00 & \cellcolor{blue!9}55.00 & \cellcolor{blue!0}52.00 & 4.00 \\
\midrule
\multirow{3}{*}{\shortstack[l]{Heterogeneous\\Qwen/Mistral/Llama}}
& MATH & \cellcolor{blue!0}44.29 & \cellcolor{blue!0}44.29 & \cellcolor{blue!3}49.29 & \cellcolor{blue!9}60.71 & \cellcolor{blue!6}57.86 & \cellcolor{blue!12}\textbf{73.57} & 29.28 \\
& BBH & \cellcolor{blue!0}56.00 & \cellcolor{blue!6}66.00 & \cellcolor{blue!3}63.00 & \cellcolor{blue!6}66.00 & \cellcolor{blue!9}69.00 & \cellcolor{blue!12}\textbf{73.00} & 17.00 \\
& MMLU-Pro & \cellcolor{blue!0}43.00 & \cellcolor{blue!3}44.00 & \cellcolor{blue!5}45.00 & \cellcolor{blue!8}47.00 & \cellcolor{blue!9}49.00 & \cellcolor{blue!12}\textbf{53.00} & 10.00 \\
\bottomrule
\end{tabular}
\end{table}

\begin{table}[htbp]
\caption{Test tasks completed by each model-backed agent in the heterogeneous configuration.
The rows correspond to the selected $V=0$ and $V=80$ runs in Table~\ref{tab:v-sensitivity}.
Each count identifies the agent responsible for the final task completion.}
\label{tab:heterogeneous-allocation}
\centering
\small
\setlength{\tabcolsep}{8pt}
\begin{tabular}{llrrr}
\toprule
Dataset & $V$ & Qwen & Mistral & Llama \\
\midrule
\multirow{2}{*}{MATH}
& 0 & 46 & 48 & 46 \\
& 80 & 138 & 0 & 2 \\
\midrule
\multirow{2}{*}{BBH}
& 0 & 37 & 30 & 33 \\
& 80 & 88 & 0 & 12 \\
\midrule
\multirow{2}{*}{MMLU-Pro}
& 0 & 31 & 34 & 35 \\
& 80 & 64 & 5 & 31 \\
\bottomrule
\end{tabular}
\end{table}

\FloatBarrier

\Needspace{0.85\textheight}
\section{Case study}
\label{app:qwen-case-study}
We follow one BBH movie-recommendation task through DEALS with three Qwen2.5-7B-Instruct agents and $V=20$.
The trace shows one split followed by one forwarding transition and execution at a second agent.

\noindent\textbf{1. Task entry.}
The task initially enters agent $a_2$ with the following question:
\begin{quote}
Find a movie similar to Pulp Fiction, The Shawshank Redemption, Dances with Wolves, Stargate:
\par
Options: (A) The Fugitive \quad (B) The Boss of It All
\par
(C) Barb Wire \quad (D) Basic Instinct 2
\end{quote}

\noindent\textbf{2. Split and partial execution at $a_2$.}
The executor returns \texttt{DECISION: split} and first summarizes the genres and themes of the four reference films.
For example, its partial output includes:
\begin{quote}
``The Shawshank Redemption is a Drama, Crime film about hope, friendship, and redemption, with deep character studies and powerful performances.''
\end{quote}
This stage generates film descriptions rather than a final choice.
The completion check returns \texttt{DECISION: Incompleted} because the four candidate movies have not yet been compared with the reference films.

\noindent\textbf{3. Requeue and forwarding.}
Next, the remaining work returns to the local queue and is forwarded from $a_2$ to $a_3$ by the DEALS router.
The continuation carries the partial result rather than restarting the task from the original question alone.
Specifically, the prompt at $a_3$ includes the film descriptions under \texttt{Completed Subtasks and Results}.

\noindent\textbf{4. Continuation and final answer at $a_3$.}
Agent $a_3$ selects \texttt{execute} and compares the candidate movies using the preceding descriptions and its retrieved demonstrations.
Its reasoning identifies \textit{The Fugitive} as the closest match through shared crime and drama elements.
The recorded final response is \texttt{RESULT: (A)}, which matches the reference answer.

Overall, the example illustrates the workflow. The task contains one split and one forward with two distinct executing agents.
Agent $a_2$ characterizes the reference films, while agent $a_3$ completes the candidate comparison.

\typeout{get arXiv to do 4 passes: Label(s) may have changed. Rerun}
\end{document}